\pdfoutput=1

\documentclass[10pt, a4paper]{article}

\usepackage[final]{lrec2026}

\usepackage{times}
\usepackage{latexsym}

\usepackage{amsmath}
\usepackage{amssymb}
\usepackage{amsfonts}
\usepackage[T1]{fontenc}

\usepackage{hyperref}
\hypersetup{
    pdfauthor={},
    pdftitle={},
    pdfsubject={},
    pdfkeywords={},
    pdfcreator={},
    pdfproducer={}
}

\usepackage[utf8]{inputenc}

\usepackage{microtype}

\usepackage{inconsolata}

\usepackage{filecontents}

\usepackage{xcolor}

\usepackage{multirow}
\usepackage{graphicx}
\usepackage{array}
\usepackage{tabularx}
\usepackage{lipsum}
\usepackage{arydshln}
\usepackage{paralist}
\usepackage{booktabs}  
\usepackage{pifont} 

\title{Disentangling Ambiguity from Instability in Large Language Models: A Clinical Text-to-SQL Case Study}

\name{Angelo Ziletti\textsuperscript{$*$}, Leonardo D'Ambrosi}

\address{Bayer AG \\
  angelo.ziletti@bayer.com}
  
\abstract{
Deploying large language models for clinical Text-to-SQL requires distinguishing two qualitatively different causes of output diversity: (i) input ambiguity that should trigger clarification, and (ii) model instability that should trigger human review. We propose CLUES, a framework that models Text-to-SQL as a two-stage process (interpretations $\rightarrow$ answers) and decomposes semantic uncertainty into an ambiguity score and an instability score. The instability score is computed via the Schur complement of a bipartite semantic graph matrix.
Across AmbigQA/SituatedQA (gold interpretations) and a clinical Text-to-SQL benchmark (known interpretations), CLUES improves failure prediction over state-of-the-art Kernel Language Entropy. In deployment settings, it remains competitive while providing a diagnostic decomposition unavailable from a single score. The resulting uncertainty regimes map to targeted interventions - query refinement for ambiguity, model improvement for instability. The high-ambiguity/high-instability regime contains 51\% of errors while covering 25\% of queries, enabling efficient triage.
 \\ \newline
\Keywords{Uncertainty Quantification, Text-to-SQL, Semantic Entropy, Clinical NLP, Electronic Health Records}
}

\begin{document}
\maketitleabstract
\section{Introduction} 
Text-to-SQL systems powered by Large Language Models (LLMs) promise to democratize access to Electronic Health Records (EHRs) and claims databases, enabling epidemiologists to query complex data using natural language~\cite{lee2024ehrsql,ziletti-dambrosi-2024-retrieval,koretsky2025biomedsql}. While retrieval-augmented approaches (RAG) have shown promising results~\cite{ziletti-dambrosi-2024-retrieval}, deployment faces a critical barrier: the risk of syntactically correct but semantically erroneous queries that return plausible yet incorrect results.

A key challenge is \textit{latent ambiguity}: ambiguity that is not evident to the user but significantly affects results~\cite{saparina2024ambrosia,saparina-lapata-2025-disambiguate}. Consider ``How many patients over 18 have atopic dermatitis?'': does ``over 18'' refer to age at diagnosis or current age? First diagnosis or any? Each interpretation yields a different SQL query and potentially different results. Left unresolved, such ambiguities undermine reproducibility of epidemiological research~\cite{zozus2016reproducibility,denaxas2017reproducibility}.

Latent ambiguity also complicates uncertainty estimation. When a model produces diverse outputs, is this diversity a legitimate reflection of query ambiguity, or a sign of model instability? 
Existing methods provide a single score that conflates two fundamentally distinct sources: uncertainty arising from inherent ambiguity in the input (akin to aleatoric uncertainty), and uncertainty stemming from the model's internal instability or lack of knowledge (akin to epistemic uncertainty). This distinction enables targeted diagnosis of system failures for model improvement. For a deployed system, it is also operationally critical: high ambiguity should trigger a clarification dialogue, while high instability should flag the query for human review.

\noindent\textbf{Contributions.}
\textbf{(1)} CLUES, a framework that decomposes semantic uncertainty into ambiguity ($H_I$) and conditional instability ($H_{R|I}$) via the Schur complement of a bipartite similarity matrix, enabling different interventions (clarification vs.\ review) and applicable to any black-box LLM.
\textbf{(2)} An interpretation-generation procedure for epidemiological questions and a clinical Text-to-SQL dataset with multiple interpretations.\footnote{\url{https://github.com/Bayer-Group/clues-clinical-nlp}}
\textbf{(3)} Empirical evidence across open-domain QA and clinical Text-to-SQL that CLUES improves failure prediction and yields regime-based error concentration useful for targeted routing.

\noindent\textbf{Paper Organization.}
Sec.~\ref{sec:methodology} formalizes our two-stage generative framework and introduces the Schur complement construction. We validate the decomposition across three settings of increasing complexity: open-domain QA with gold interpretations (Sec.~\ref{sec:experiments_open_datasets}), a real-world clinical Text-to-SQL dataset (Sec.~\ref{sec:clinical_application_known_interpretations}), and production deployment with on-the-fly interpretation generation (Sec.~\ref{sec:clinical_application_deployment}).

\section{Related Work} 
\label{sec:related_work}

\noindent\textbf{Semantic entropy and kernel-based uncertainty.}
Semantic entropy clusters generations into equivalence classes for hallucination detection\cite{wiegreffe2023measuring,manakul2024detecting}; Kernel Language Entropy (KLE) generalizes this by encoding pairwise semantic similarities.\cite{nikitin2024kernel} Recent work derives kernel entropy via bias-variance-covariance decomposition,\cite{gruber2024biasvariance} combines semantic signals with token-level uncertainty,\cite{raghuvanshi2025} and exploits geometric measures such as semantic density and volume.\cite{zhaoye2025semanticdensity,li2025semanticvolume,grewal2024improving} However, these methods operate on a single pool of outputs and do not model multi-stage generative processes or distinguish uncertainty sources.\cite{nikitin2024kernel,gruber2024biasvariance}

\noindent\textbf{Decomposing aleatoric and epistemic uncertainty.}
This distinction is fundamental in probabilistic machine learning and critical for deployment in safety-critical domains.\cite{kendall2017uncertainties,hullermeier2021aleatoric} Recent LLM work trains meta-models or uses temperature sensitivity to identify epistemic uncertainty, but these approaches require additional supervision or pipeline changes.\cite{desai2024distinguishing,foodeei2025semantic,song2025inventropy} In the EHR-QA domain, \citet{kim2022uncertainty} decompose uncertainty to detect ambiguous questions, but operate at the token level without modeling multi-stage generation.

\noindent\textbf{Text-to-SQL and clinical uncertainty.} 
Text-to-SQL has evolved from cross-domain benchmarks~\cite{yu-etal-2018-spider} to clinical applications for EHR access and clinical trial recruitment~\cite{ziletti2025generatingpatientcohortselectronic,ghosh-etal-2025-survey-llm,deng-etal-2022-recent}. Benchmarks like EHRSQL~\cite{lee2024ehrsql} and BiomedSQL~\cite{koretsky2025biomedsql} evaluate clinical Text-to-SQL, while multi-turn approaches improve robustness~\cite{ryu-etal-2024-ehr}. Existing systems prioritize execution accuracy but provide limited uncertainty quantification~\cite{kim2022uncertainty,jo-etal-2024-lg,ziletti-dambrosi-2024-retrieval}. Reliability work focuses on unanswerable detection, ensemble methods, or token entropy thresholds~\cite{jo-etal-2024-lg,kim-etal-2024-probgate}, but does not decompose uncertainty by source or model semantic equivalence.

\section{Methodology: A Decomposition Framework for Semantic Uncertainty} \label{sec:methodology}
\begin{figure*}[t!]
\centering
\includegraphics[width=\textwidth]{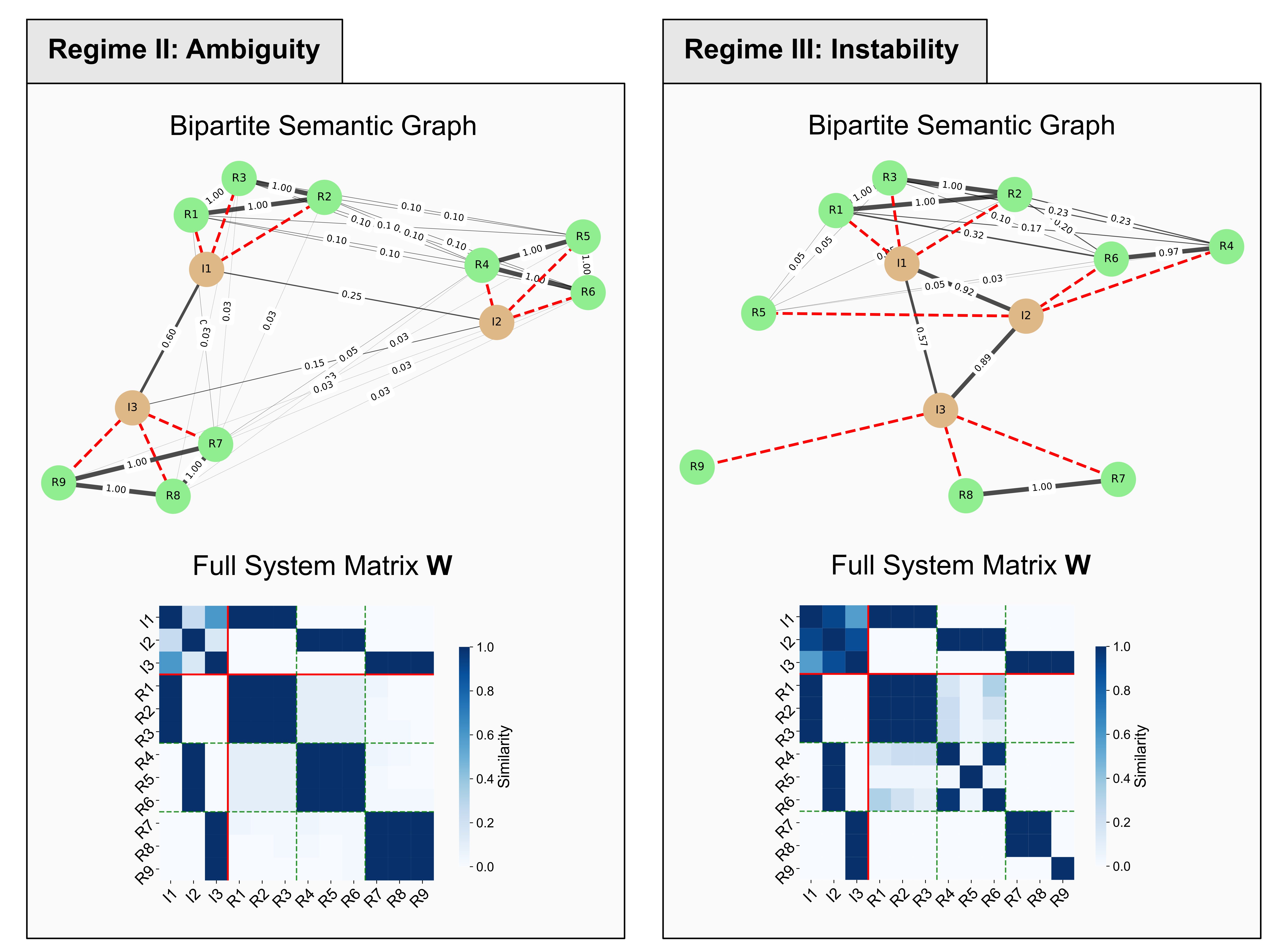}
\caption{Examples of uncertainty regimes in CLUES. (Left) Regime II, Ambiguity: Interpretations (brown) are semantically distinct (low $\mathbf{W}_{II}$ off-diagonal), but results (green) cluster tightly within each interpretation (high intra-cluster $\mathbf{W}_{RR}$), yielding high $H_I$, low $H_{R|I}$. (Right) Regime III, Instability: Interpretations are similar (high $\mathbf{W}_{II}$ off-diagonal), yet results vary substantially within interpretations, yielding low $H_I$, high $H_{R|I}$. Red dashed edges: interpretation-result assignments ($\mathbf{W}_{IR}$). Bottom: corresponding system matrices $\mathbf{W}$.}
\label{fig:case-study-visuals}
\end{figure*}
Our methodology extends the geometric view of uncertainty from KLE to a multi-stage generative process. We construct a bipartite semantic graph representing the entire system and use tools from linear algebra and spectral graph theory to decompose its structural complexity.

\subsection{Interpretation-Augmented Generation}
We formalize the Text-to-SQL system as a two-stage generative process. Given an initial natural language query $q$:

\begin{enumerate}
    \item \textbf{Interpretation Stage:} The system first generates a set of $N$ distinct semantic interpretations, $\mathcal{I} = \{I_1, I_2, \dots, I_N\}$. Each interpretation $I_n$ is a natural language reformulation representing one plausible reading of the original query $q$. This stage explicitly surfaces the ambiguity (or lack thereof) inherent in the input.
    
    \item \textbf{Generation Stage:} For each interpretation $I_n \in \mathcal{I}$, the system generates $M$ candidate SQL queries, executes them against the database, and verbalizes the retrieved data into natural language answers. This yields a set of answers $\mathcal{R}_n = \{R_{n,1}, R_{n,2}, \dots, R_{n,M}\}$ per interpretation. The complete answer set is $\mathcal{R} = \{R_{1,1}, \dots, R_{N,M}\}$, containing $N \times M$ total answers.
\end{enumerate}

This one-to-many ($N$ $\rightarrow$ $N \times M$) structure allows us to distinguish diversity \textit{across} interpretations (ambiguity) from instability \textit{within} the generation process for a single interpretation.

\subsection{Constructing the Bipartite Semantic Graph}

We define a similarity function $k(\cdot, \cdot) \in [0, 1]$ measuring semantic similarity between text strings. The full system matrix $\mathbf{W}$ serves as the graph's weighted adjacency matrix and has a natural bipartite block structure:
\begin{equation}
\label{eq:block_matrix}
\mathbf{W} =
\begin{pmatrix}
\mathbf{W}_{II} & \mathbf{W}_{IR} \\
\mathbf{W}_{RI} & \mathbf{W}_{RR}
\end{pmatrix}
\in \mathbb{R}^{(N+NM) \times (N+NM)}
\end{equation}
where:
\begin{itemize}
    \item $\mathbf{W}_{II} \in \mathbb{R}^{N \times N}$: The \emph{interpretation similarity matrix}, with $(\mathbf{W}_{II})_{ij} = k(I_i, I_j)$. This block encodes the semantic relationships among interpretations.
    \item $\mathbf{W}_{RR} \in \mathbb{R}^{NM \times NM}$: The \emph{answer similarity matrix} (flattening $(n,m)\mapsto j$), with $(\mathbf{W}_{RR})_{ij} = k(R_i, R_j)$. This block encodes the semantic relationships among all generated answers.
    \item $\mathbf{W}_{IR} \in \mathbb{R}^{N \times NM}$ and $\mathbf{W}_{RI} = \mathbf{W}_{IR}^\top$: The \emph{assignment matrices}, with $(\mathbf{W}_{IR})_{ij} = 1$ if answer $R_j$ was generated from interpretation $I_i$, and $0$ otherwise.
\end{itemize}

The diagonal blocks $\mathbf{W}_{II}$ and $\mathbf{W}_{RR}$ encode graded semantic similarity, interpreted as weighted edge connectivity in the graph. The off-diagonal blocks $\mathbf{W}_{IR}$ and $\mathbf{W}_{RI}$ encode the generative provenance between interpretations and their answers. We represent this provenance as binary connectivity, since each answer traces to exactly one interpretation. This mixed representation is consistent with the graph-theoretic framework: edge weights encode connection strength, whether derived from continuous similarity scores or binary structural links. The KLE framework \cite{nikitin2024kernel} constructs graphs from natural-language-inference-based similarity scores; our extension introduces a second node type (answers) connected to the first (interpretations) via known generative links rather than inferred similarity.

\noindent\textbf{Prompt-Based Similarity.}
The similarity function $k(\cdot, \cdot)$ is implemented via an LLM with a task-specific prompt that defines the relevant notion of equivalence. For Text-to-SQL, the prompt assesses whether two interpretations would yield logically equivalent SQL queries; for QA benchmarks, whether they would yield the same factual answer. This ensures that $\mathbf{W}_{II}$ and $\mathbf{W}_{RR}$ encode task-relevant semantics rather than generic lexical similarity. 

\noindent\textbf{Heat Kernel Regularization.}
Following KLE~\cite{nikitin2024kernel}, we apply a heat diffusion kernel $\mathbf{K}_{\tau} = e^{-\tau\mathbf{L}}$, where $\mathbf{L} = \mathbf{D} - \mathbf{W}$ is the graph Laplacian and $\mathbf{D}$ is the degree matrix. This transforms raw pairwise similarities into a smoothed representation where the hyperparameter $\tau$ controls granularity: small $\tau$ preserves fine-grained distinctions, while large $\tau$ diffuses similarity across the graph, merging nearby nodes into coarse clusters.
To select $\tau$ without a validation set, we calibrate against an idealized baseline: $\mathbf{W}_{RR}$ of perfectly identical items (all ones) should yield near-zero entropy ($<$0.001 bits). This yields $\tau = 10$ for our experiments. Sensitivity analysis (Table~\ref{tab:tau_sensitivity}) confirms robustness across $\tau \in [2, 20]$.

\subsection{Decomposing Uncertainty with the Schur Complement}
\label{sec:decomposition}
Our central goal is to decompose the total uncertainty of the system into interpretable components: \textit{input ambiguity} and \textit{system instability}. The resulting scores $H_I$ and $H_{R|I}$ are not intended to replace aggregate measures like $H_R$ for failure prediction; rather, they provide complementary information about the \emph{source} of uncertainty, enabling differential interventions (input refinement vs.\ generation review) that a single score cannot support.

\subsubsection{Baseline: Result Entropy $H_R$}
The state-of-the-art KLE approach~\cite{nikitin2024kernel} computes entropy over generated outputs without modeling their provenance. In our framework, this yields $H_R$ from $\mathbf{W}_{RR}$ alone.
High $H_R$ indicates diverse outputs but does not distinguish the source of uncertainty: input ambiguity or system instability. This is the baseline we aim to improve upon.

\subsubsection{Joint Uncertainty $H(R,I)$}
We define total system uncertainty as the joint entropy over the full bipartite graph, computed from the complete similarity matrix $\mathbf{W}$ (Eq.~\ref{eq:block_matrix}). This captures uncertainty across both interpretations and results, along with their generative relationships encoded in $\mathbf{W}_{IR}$.

\subsubsection{Ambiguity Score $H_I$}
We define the ambiguity score as the entropy over the set of interpretations, estimating ambiguity in the user's query. We compute $H_I$ by applying the KLE framework to $\mathbf{W}_{II}$:
\begin{equation}
\begin{aligned}
\label{eq:hi_definition}
\mathbf{K}_I &= e^{-\tau\mathbf{L}_I}, \quad \boldsymbol{\rho}_I = \mathbf{K}_I / \text{Tr}(\mathbf{K}_I), \\
H_I &= -\text{Tr}(\boldsymbol{\rho}_I \log \boldsymbol{\rho}_I)
\end{aligned}
\end{equation}

\subsubsection{Instability Score $H_{R|I}$}
We define the instability score as the conditional entropy of results given interpretations: ``Given a clear interpretation, how much does the answer still vary?''

\noindent\textbf{The Naive Subtraction Approach.}
A natural approach is to estimate conditional entropy via subtraction: $H_{R|I} = H_{R,I} - H_I$. However, this relies on the Shannon entropy chain rule, which does not generally hold for von Neumann entropy on heat-kernel density matrices\cite{nikitin2024kernel}. The joint and marginal kernels have different sizes (see Eq.~\ref{eq:block_matrix}) and undergo independent diffusion processes, so $\tilde{H}_{R|I}$ is not guaranteed to be positive. Indeed, $\tilde{H}$ yields negative values in 37--60\% of cases (Tables~\ref{tab:multimodel_open_qa} and~\ref{tab:ehr_results_0.6}).

\noindent\textbf{The Schur Complement Approach.}
We leverage the block structure of $\mathbf{W}$ (Eq.~\ref{eq:block_matrix}) to construct the conditional similarity structure directly via the Schur complement~\cite{Schur1917-zo,Zhang2005-dc}:
\begin{equation}
\label{eq:schur_complement}
\mathbf{S} = \mathbf{W}_{RR} - \mathbf{W}_{RI} (\mathbf{W}_{II} + \epsilon\mathbf{I})^{-1} \mathbf{W}_{IR}
\end{equation}
where $\epsilon = 10^{-3}$ ensures invertibility.
The Schur complement provides a principled way to ``condition on'' the interpretation structure. The projection term $\mathbf{W}_{RI} \mathbf{W}_{II}^{-1} \mathbf{W}_{IR}$ captures the portion of result similarity explained by interpretations. The residual $\mathbf{S}$ retains only the similarity structure that input ambiguity cannot account for. This mirrors the Gaussian case, where the Schur complement yields conditional covariance $\text{Cov}(R|I)$~\cite{Boyd2004-mw}. The entropy $H_{R|I}$ computed from $\mathbf{S}$ therefore quantifies the system's internal inconsistency.
Since $\mathbf{S}$ is not guaranteed to be positive semidefinite (PSD), we project onto the PSD cone via eigendecomposition, clipping negative eigenvalues to zero~\cite{higham2002}, then apply the KLE recipe (Eq.~\ref{eq:hi_definition}). Unlike the subtraction approach above, this guarantees $H_{R|I} \geq 0$.
We term this decomposition framework \textbf{CLUES} (Conditional Language Uncertainty via Entropy and Schur). Crucially, CLUES distinguishes uncertainty from query ambiguity (potentially addressable via clarification) from system instability (requiring model-level intervention).

\subsubsection{Uncertainty regimes and recommended interventions}
\label{sec:regimes_definition}

We define four regimes by thresholding $H_I$ (ambiguity) and $H_{R|I}$ (instability):
\begin{itemize}
  \item Regime I: Confident (low $H_I$, low $H_{R|I}$): auto-answer.
  \item Regime II: Ambiguity (high $H_I$, low $H_{R|I}$): input refinement (clarification).
  \item Regime III: Instability (low $H_I$, high $H_{R|I}$): generation review.
  \item Regime IV: Compound (high/high): clarification + generation review.
\end{itemize}
In experiments we use median thresholds unless otherwise stated.
We evaluate both failure prediction and regime-based routing in Sec.~\ref{sec:ambiqa_situatedqa_results}, \ref{subsec:ehr_results}, and \ref{sec:regime_analysis}.

\section{Experiments on Open Answer Datasets}
\label{sec:experiments_open_datasets}

\begin{table*}[h!tbp]
\centering
\resizebox{\textwidth}{!}{%
\begin{tabular}{l c ccc cc c | c ccc cc c}
\toprule
& \multicolumn{7}{c|}{\textbf{AmbigQA} ($N_q=300$)} & \multicolumn{7}{c}{\textbf{SituatedQA} ($N_q=300$)} \\
\cmidrule(lr){2-8} \cmidrule(lr){9-15}
& & \multicolumn{3}{c}{AUROC} & \multicolumn{2}{c}{Regime Acc.} & & & \multicolumn{3}{c}{AUROC} & \multicolumn{2}{c}{Regime Acc.} & \\
\cmidrule(lr){3-5} \cmidrule(lr){6-7} \cmidrule(lr){10-12} \cmidrule(lr){13-14}
\textbf{Model} & \textbf{Acc.} & $H_R$ & $\tilde{H}_{R|I}$ & $H_{R|I}$ & \textbf{A} & \textbf{B} & $\Delta$ & \textbf{Acc.} & $H_R$ & $\tilde{H}_{R|I}$ & $H_{R|I}$ & \textbf{A} & \textbf{B} & $\Delta$ \\
\midrule
Qwen3 & 25.0 & .476 & \textbf{.550} & .506 & 29.8 & 23.7 & +6.2 & 20.7 & .494 & .521 & \textbf{.561} & 32.8 & 16.3 & +16.5 \\
GPT-OSS & 27.0 & .641 & .524 & \textbf{.702} & 53.8 & 8.1 & +45.7 & 25.7 & .660 & .634 & \textbf{.758} & 44.4 & 3.1 & +41.3 \\
KimiK2 & 40.7 & .663 & .466 & \textbf{.770} & 76.1 & 16.3 & +59.7 & 37.3 & .631 & .590 & \textbf{.752} & 64.3 & 13.8 & +50.5 \\
Claude4.5S & 35.7 & .501 & .475 & \textbf{.533} & 40.7 & 35.7 & +5.0 & 30.7 & .502 & \textbf{.569} & .495 & 44.4 & 19.5 & +24.9 \\
Gemini3Pro & 47.3 & .538 & .480 & \textbf{.659} & 73.8 & 32.6 & +41.2 & 44.6 & .584 & .594 & \textbf{.685} & 61.6 & 12.7 & +48.9 \\
\midrule
\textit{Pooled} & \textit{35.1} & \textit{.550} & \textit{.498} & \textit{\textbf{.627}} & \textit{56.6} & \textit{21.1} & \textit{+35.5} & \textit{31.8} & \textit{.567} & \textit{.573} & \textit{\textbf{.641}} & \textit{46.8} & \textit{13.3} & \textit{+33.5} \\
\bottomrule
\end{tabular}%
}
\caption{Multi-Model Validation on Open-Domain QA Benchmarks. We evaluate $N_q=300$ questions per dataset across 5 LLMs (pooled: $N_q \times 5 = 1500$ question--model instances). We compare $H_R$ (baseline KLE result entropy), naive subtraction $\tilde{H}_{R|I}$, and Schur-complement conditional entropy $H_{R|I}$ (CLUES) for failure prediction (AUROC). $\Delta$ denotes the accuracy gap between Regime A (high $H_R$, low $H_{R|I}$) and Regime B (high $H_R$, high $H_{R|I}$), where high/low are defined by pooled median splits. The naive subtraction $\tilde{H}_{R|I}$ yields negative values in 53\%/37\% of cases (AmbigQA/SituatedQA).}
\label{tab:multimodel_open_qa}
\end{table*}

We first validate our decomposition on two open-domain QA benchmarks: AmbigQA~\cite{min-etal-2020-ambigqa}, where questions admit multiple valid interpretations due to semantic ambiguity, and SituatedQA~\cite{zhang-choi-2021-situatedqa}, where answers vary based on temporal or geographical context. Both datasets provide gold-standard interpretations paired with corresponding answers, allowing us to test whether $H_{R|I}$ provides a more accurate failure signal than $H_R$.

\subsection{Experimental Setup}
\label{subsec:dataset_protocol_ambiqa_situatedqa}

\noindent\textbf{Dataset and Protocol.}
From each dataset, we select the first $N_q=300$ questions with at least two gold interpretations, capping at three interpretations per question to limit inference costs.
Following the two-stage framework in Sec.~\ref{sec:methodology}, the interpretation stage is given by the dataset's gold interpretations. For the generation stage, we sample $M=3$ answers per interpretation, yielding $N \times M$ results per question ($N \in \{2, 3\}$). We construct the bipartite semantic graph $\mathbf{W}$ and compute $H_R$, $\tilde{H}_{R|I}$, and $H_{R|I}$.

\noindent\textbf{Large Language Models.}
For answer generation, we evaluate five frontier LLMs: GPT-OSS-120B (hereafter \textit{GPT-OSS})~\cite{openai2025gptoss120bgptoss20bmodel}, Kimi K2 Thinking (\textit{KimiK2})~\cite{kimiteam2025kimik2openagentic} and Qwen-3-VL-235B-A22B (\textit{Qwen3})~\cite{qwen3technicalreport} are open-weights; Gemini 3 Pro (\textit{Gemini3Pro})~\cite{geminiteam2023gemini} and Claude Sonnet 4.5 (\textit{Claude4.5S})~\cite{anthropic2023claude} are proprietary. All models are sampled with temperature $T=1$. Throughout this paper, we use Gemini 2.5 Flash ($T=0$) for semantic similarity ($\mathbf{W}_{II}$, $\mathbf{W}_{RR}$) and answer correctness evaluation (unless otherwise noted). 

\noindent\textbf{Evaluation Metric.}
We employ a strict path consistency strategy: a generated answer is correct only if it matches the gold answer for the specific interpretation that produced it, not any valid answer. Correctness is assessed via LLM-as-a-judge, prompting Gemini3Pro with the predicted and gold answers. A question is labeled as failure if the proportion of correct answers falls below a threshold $\eta$ (we use $\eta=0.8$). This criterion requires the system to respect the semantic constraints of each interpretation, penalizing both mode collapse (identical outputs across distinct interpretations) and context confusion (correct answer paired with wrong interpretation).

\subsection{Results and Analysis}
\label{sec:ambiqa_situatedqa_results}

Results are presented in Table~\ref{tab:multimodel_open_qa}. Predicting failures on frontier LLMs is inherently challenging; the baseline $H_R$ achieves AUROC scores of 0.47--0.66 depending on model and dataset. Our proposed $H_{R|I}$ consistently outperforms $H_R$, achieving pooled AUROC of 0.627 on AmbigQA (+0.077) and 0.641 on SituatedQA (+0.074). Per-model results show consistent improvements, with the largest gains when $H_R$ itself carries predictive signal; for models where $H_R$ performs near chance (Claude4.5S, Qwen3), all entropy measures show very limited discriminative power.
The naive subtraction approach $\tilde{H}_{R|I}$ achieves near-chance AUROC, validating the need for the Schur complement construction (Eq.\ref{eq:schur_complement}).

The improvement over $H_R$ stems from CLUES's ability to account for input structure. By ignoring interpretations, $H_R$ errs in both directions: \textit{diversity inflation}, where legitimate variation across interpretations ($\mathbf{W}_{II} \approx \mathbf{I}$, $\mathbf{W}_{RR} \approx \mathbf{I}$) is flagged as uncertainty; and \textit{mode collapse}, where identical outputs across distinct interpretations ($\mathbf{W}_{RR} \approx \mathbf{1}$) yield false confidence. The Schur complement detects both failure modes.

\begin{table*}[ht!]
\centering
\small
\begin{tabularx}{\textwidth}{>{\raggedright\arraybackslash}p{0.32\textwidth} >{\raggedright\arraybackslash}X}
\toprule
\textbf{Input Question} & \textbf{Disambiguated Question} \\
\midrule
How many patients > 17 yo have atopic dermatitis (all codes). Breakdown by code & 
Calculate the count of \textbf{unique patients} who have a \textbf{first diagnosis} of Atopic Dermatitis at any time in their patient history, where the patient's \textbf{age at the time of this first diagnosis} was greater than 17 years. Provide this count broken down by the specific concept code of Atopic Dermatitis. \\
\midrule
How many patients with chronic kidney disease never took heparin before their chronic kidney disease diagnosis? & 
Count the number of unique patients who have a \textbf{first occurrence} of Chronic Kidney Disease and who have \textbf{no record} of Heparin administration at any time \textbf{prior to their first} Chronic Kidney Disease diagnosis. \\
\bottomrule
\end{tabularx}
\caption{Disambiguation examples resolving temporal, demographic, and event ordering ambiguities. Bold text indicates specifications that were implicit or absent in the original question.}
\label{tab:disambiguation_examples}
\end{table*}

\paragraph{Regime Analysis.}
The four-regime framework (Sec.~\ref{sec:regimes_definition}) partitions by $H_I$ and $H_{R|I}$ to guide interventions. Here, we test whether $H_{R|I}$ provides discriminative value beyond $H_R$ alone. We focus on high-uncertainty queries where $H_R$ exceeds its median, then partition by median $H_{R|I}$ into Regime~A (low $H_{R|I}$) and Regime~B (high $H_{R|I}$). Despite identical $H_R$ profiles, accuracy differs substantially: 56.6\% vs.\ 21.1\% for AmbigQA and 46.8\% vs.\ 13.3\% for SituatedQA. These differences are highly significant (Chi-square $p < 10^{-22}$), with odds ratios of 4.9 and 5.7: queries in Regime~B are 5--6$\times$ more likely to fail. This pattern holds across all five LLMs without model-specific calibration.

\section{Clinical Text-to-SQL with Known Interpretations}
\label{sec:clinical_application_known_interpretations}

\subsection{Resolving Ambiguity in Epidemiological Questions}
\label{sec:disambiguation}
Epidemiological questions in natural language frequently contain implicit ambiguities requiring explicit specification for accurate analysis (Table~\ref{tab:disambiguation_examples}).
While rule-based templates (e.g., OHDSI ATLAS~\cite{OHDSIAtlas}) lack flexibility and interactive clarification introduces user fatigue, our iterative disambiguation approach automatically infers the most plausible interpretation through successive refinement rounds, presenting users with a transparent final specification they can verify or edit before SQL generation.
The disambiguation pipeline operates iteratively until convergence. Given an input question, each round prompts the LLM to:
\begin{enumerate}
    \item Identify common epidemiological ambiguities including patient count semantics (unique patients vs.\ all records), temporal relationships (before/after, within timeframes), population definitions (inclusion/exclusion criteria), event ordering (first vs.\ any occurrence), and demographic specifications (age at diagnosis vs.\ current age).
    \item Assign an ambiguity score $s \in [0,1]$ and generate multiple candidate interpretations ordered by clinical plausibility.
    \item Select the most plausible interpretation for the next round.
\end{enumerate}

\noindent The prompt embeds domain conventions as defaults reflecting common practice in observational health research (e.g., standard age bands; defaulting to entire patient history when unspecified; unique patient counts). The process terminates when ambiguity falls below a threshold ($s < 0.1$); in practice, within 1--3 iterations. We use Gemini 2.5 Flash with structured output schemas.

This iterative disambiguation offers three advantages: (1) transparency, as users can review and edit candidates before SQL generation; (2) educational value, as outputs serve as templates for well-formed questions; and (3) systematic application of epidemiological best practices without user-specified routine parameters.

In qualitative evaluation, epidemiologists confirmed that disambiguated outputs accurately reflect domain conventions and produce clinically appropriate disambiguations.

\subsection{Experimental Setup}
\label{subsec:dataset_fixed_interpretations}

\noindent\textbf{Source Dataset and Interpretation Construction.}
We build on EpiAskKB~\cite{ziletti2025generatingpatientcohortselectronic}, which contains question-SQL pairs for epidemiological research on EHR. Questions may be ambiguous, reflecting how epidemiologists naturally pose queries (Table~\ref{tab:disambiguation_examples}).
For each question, we construct a multi-interpretation setup following Sec.~\ref{subsec:dataset_protocol_ambiqa_situatedqa}. From the question and gold SQL, we generate a disambiguated question reflecting the gold SQL, plus three alternative interpretations (Sec.~\ref{sec:disambiguation}). For each alternative, we generate SQL by prompting the LLM with the original question, alternative interpretation, and gold SQL to guide query structure. This constrained context minimizes structural errors. We use Gemini3Pro for all generation.

\noindent\textbf{Generation and Execution Pipeline.}
For each question, we sample $M=3$ SQL queries per interpretation, yielding $3 \times 3 = 9$ generated queries per model-temperature configuration. SQL generation, entity resolution, self-correction, and execution follow~\citet{ziletti-dambrosi-2024-retrieval}. Queries are run on Optum's de-identified Clinformatics\textsuperscript{\textregistered} Data Mart Database, and retrieved results are verbalized into natural language answers, from which we compute $H_R$, $\tilde{H}_{R|I}$, and $H_{R|I}$ as in Sec.~\ref{sec:methodology}. Evaluation follows Sec.~\ref{subsec:dataset_protocol_ambiqa_situatedqa} with $\eta = 0.6$.

\subsection{Results}
\label{subsec:ehr_results}
Results are presented in Table~\ref{tab:ehr_results_0.6}. $H_{R|I}$ outperforms $H_R$ as a failure predictor in 17 of 20 model-temperature configurations, achieving pooled AUROC of 0.762 [95\% CI: 0.737, 0.787] versus 0.600 [0.575, 0.627] for $H_R$. This difference is significant ($p < 10^{-10}$), demonstrating that conditioning on interpretation structure provides substantial discriminative value. The naive subtraction $\tilde{H}_{R|I}$ achieves only 0.461 AUROC with 60\% negative values, validating the theoretical necessity of the Schur complement (Sec.~\ref{sec:decomposition}).

\begin{table}[htbp]
\centering
\small
\resizebox{\columnwidth}{!}{%
\begin{tabular}{l c ccc c}
\toprule
& & \multicolumn{3}{c}{AUROC} & \\
\cmidrule(lr){3-5}
\textbf{Model (T)} & \textbf{Acc.} & $H_R$ & $\tilde{H}_{R|I}$ & $H_{R|I}$ & $\Delta$ \\
\midrule
\multicolumn{6}{l}{\textit{Qwen3}} \\
\quad T=0.7 & 53.3 & .605 & .544 & \textbf{.764} & +54.3 \\
\quad T=1.0 & 54.2 & .617 & .531 & \textbf{.748} & +42.6 \\
\quad T=1.5 & 55.0 & .618 & .505 & \textbf{.818} & +59.2 \\
\quad T=2.0 & 54.1 & .630 & .498 & \textbf{.786} & +65.8 \\
\midrule
\multicolumn{6}{l}{\textit{GPT-OSS}} \\
\quad T=0.7 & 71.6 & .600 & .416 & \textbf{.731} & +31.2 \\
\quad T=1.0 & 75.5 & .615 & .410 & \textbf{.828} & +34.3 \\
\quad T=1.5 & 71.6 & .615 & .384 & \textbf{.831} & +34.2 \\
\quad T=2.0 & 71.6 & .612 & .391 & \textbf{.818} & +35.0 \\
\midrule
\multicolumn{6}{l}{\textit{KimiK2 }} \\
\quad T=0.7 & 76.1 & .632 & .446 & \textbf{.815} & +41.7 \\
\quad T=1.0 & 74.3 & .548 & .417 & \textbf{.817} & +40.5 \\
\quad T=1.5 & 75.2 & .635 & .444 & \textbf{.791} & +44.7 \\
\quad T=2.0 & 78.9 & .599 & .418 & \textbf{.853} & +27.8 \\
\midrule
\multicolumn{6}{l}{\textit{Claude4.5S}} \\
\quad T=0.7 & 78.9 & .593 & .395 & \textbf{.611} & +0.0 \\
\quad T=1.0 & 77.6 & .595 & .466 & \textbf{.694} & +10.3 \\
\quad T=1.5 & 78.0 & .633 & .400 & \textbf{.744} & +30.8 \\
\quad T=2.0 & 78.9 & .575 & .426 & \textbf{.622} & +16.8 \\
\midrule
\multicolumn{6}{l}{\textit{Gemini3Pro}} \\
\quad T=0.7 & 93.3 & \textbf{.708} & .623 & .604 & +3.8 \\
\quad T=1.0 & 93.3 & .620 & .532 & \textbf{.715} & +20.6 \\
\quad T=1.5 & 88.8 & \textbf{.642} & .578 & .593 & +5.9 \\
\quad T=2.0 & 88.8 & \textbf{.635} & .542 & .595 & +10.7 \\
\midrule
\textit{Pooled} & \textit{74.4} & \textit{.600} & \textit{.461} & \textit{\textbf{.762}} & \textit{+33.6} \\
\bottomrule
\end{tabular}
}
\caption{Clinical Text-to-SQL benchmark results ($N=2{,}180$; 109 questions $\times$ 5 models $\times$ 4 temperatures). $\Delta$ = Regime A $-$ Regime B accuracy gap. $\tilde{H}_{R|I}$ produces negative values in 60\% of cases.}
\label{tab:ehr_results_0.6}
\end{table}

\noindent\textbf{Regime Analysis Confirms Discriminative Value.}
The accuracy gap $\Delta$ between Regime A (high $H_R$, low $H_{R|I}$) and Regime B (high $H_R$, high $H_{R|I}$), split at median values, is positive in 19 of 20 configurations (pooled: +33.6 pp). This replicates the open-domain QA pattern (Sec.~\ref{sec:ambiqa_situatedqa_results}): among queries with equally high output diversity, those with low $H_{R|I}$ are substantially more likely to succeed.

\noindent\textbf{Temperature Modulates Uncertainty Signal.}
For most models, moderate temperatures ($T=1.0$--$1.5$) yield the strongest $H_{R|I}$ signal. Claude 4.5 illustrates this pattern: regime separation increases from $\Delta = 0$ at T=0.7 to $\Delta = +30.8$ at T=1.5 ($0 \rightarrow +30.8$), suggesting higher sampling diversity exposes latent instability. Gemini 3 Pro is the exception: its errors appear to stem from consistent but incorrect outputs rather than instability.

\noindent\textbf{Sensitivity to Diffusion Parameter $\tau$.}
Table~\ref{tab:tau_sensitivity} shows $H_{R|I}$ is robust for $\tau \geq 2$ (AUROC 0.75--0.76), while $H_R$ degrades at high $\tau$ (0.74 $\rightarrow$ 0.58) as diffusion blurs pairwise similarities toward uniformity. The Schur complement resists this effect: both raw similarities and the projection term smooth proportionally, preserving residual signal.

\begin{table}[h!tbp]
\centering
\setlength{\tabcolsep}{4pt}
\begin{tabular}{@{}l ccccccc@{}}
\toprule
$\tau$ & 1 & 2 & 3 & 5 & 10 & 15 & 20 \\
\midrule
$H_I$     & .571 & .574 & .576 & .578 & .572 & .569 & .567 \\
$H_R$     & .669 & .742 & .738 & .676 & .601 & .586 & .582 \\
$H_{R|I}$ & .675 & .746 & .764 & .765 & .762 & .761 & .760 \\
\bottomrule
\end{tabular}
\caption{Pooled AUROC sensitivity to heat kernel parameter $\tau$. $H_{R|I}$ remains stable across $\tau \in [2, 20]$ while $H_R$ degrades at high diffusion.}
\label{tab:tau_sensitivity}
\end{table}

\section{Clinical Text-to-SQL in Deployment Settings}
\label{sec:clinical_application_deployment}

\subsection{Experimental Setup}
\label{subsec:clinical_application_deployment_proxy}

Having established that $H_{R|I}$ outperforms $H_R$ with known interpretations, we now test whether these gains extend to production deployment. This setting differs from previous experiments in three key ways.

\noindent\textbf{Interpretations are generated, not given.} Interpretations are produced on the fly via our disambiguation procedure (Sec.~\ref{sec:disambiguation}) and may imperfectly capture the true ambiguity structure, unlike the validated gold/silver labels used previously.

\noindent\textbf{Uncertainty serves as a proxy signal.} We evaluate final-answer correctness only, meaning $H_{R|I}$ computed over interpretations serves as a proxy for reliability rather than directly measuring per-interpretation accuracy.

\noindent\textbf{Limited statistical power.} Evaluation is coarser (one label per question rather than per interpretation), and models achieve high accuracy (Table~\ref{tab:model_accuracy}), yielding sparse errors that challenge statistical estimation.

We use the same pipeline as Sec.~\ref{subsec:dataset_fixed_interpretations}, testing with and without disambiguation and RAG. RAG follows the methodology in~\citet{ziletti-dambrosi-2024-retrieval}.

\subsection{Results}
Table~\ref{tab:model_accuracy} summarizes performance in the deployment setting. All models achieve 84--100\% accuracy, higher than Sec.~\ref{subsec:ehr_results}. This reflects the evaluation setup: strict path consistency requires correctness across multiple interpretations including less natural ones, while here we evaluate a single answer against the original query, which typically matches the model's default interpretation.

\begin{table}[h!tbp]
\centering
\setlength{\tabcolsep}{3pt}
\begin{tabular}{@{}cc ccccc c@{}}
\toprule
& & \rotatebox{90}{\textbf{Qwen3}} & \rotatebox{90}{\textbf{GPT-OSS}} & \rotatebox{90}{\textbf{KimiK2}} & \rotatebox{90}{\textbf{Claude4.5S}} & \rotatebox{90}{\textbf{Gemini3Pro}} & \rotatebox{90}{\textbf{Pooled}} \\
\textbf{Dis} & \textbf{RAG} & & & & & & \\
\midrule
$\times$ & $\times$ & 85.3 & 93.6 & 99.1 & 95.4 & 100 & 94.7 \\
$\checkmark$ & $\times$ & 88.1 & 98.2 & 99.1 & 94.5 & 97.2 & 95.4 \\
$\times$ & $\checkmark$ & 90.8 & 99.1 & 97.2 & 97.2 & 100 & 96.9 \\
$\checkmark$ & $\checkmark$ & 94.5 & 92.7 & 94.5 & 96.3 & 100 & 95.6 \\
\midrule
\multicolumn{2}{@{}l}{\textbf{Pooled}} & 89.7 & 95.9 & 97.5 & 95.9 & 99.3 & 95.6 \\
\bottomrule
\end{tabular}
\caption{Model Accuracy by Configuration. Dis.\ = disambiguated input. All values in \%.}
\label{tab:model_accuracy}
\end{table}

\noindent\textbf{Uncertainty Separates Correct from Incorrect.}
Despite sparse errors, median uncertainty values are systematically higher for incorrect predictions. Pooled across all configurations ($N=2{,}180$), incorrect predictions show median $H_I$ of 0.189 vs.\ 0.069 for correct, $H_R$ of 0.442 vs.\ 0.301, and $H_{R|I}$ of 0.072 vs.\ 0.003.

\noindent\textbf{RAG Stabilizes Output.}
RAG substantially reduces $H_{R|I}$ for correct predictions: GPT-OSS drops from 0.113 to 0.030, Qwen3 from 0.046 to 0.006, and KimiK2 from 0.062 to 0.025. However, RAG also reduces $H_{R|I}$ for incorrect predictions (pooled median: 0.104 $\rightarrow$ 0.052), which may suppress the uncertainty signal that would otherwise flag errors (e.g., GPT-OSS: 0.296 $\rightarrow$ 0.025). Claude4.5S and Gemini3Pro show minimal $H_{R|I}$ (0.002) with or without RAG, suggesting sufficient internalized domain knowledge.

\noindent\textbf{AUROC and regime-based diagnostics.}
In this setting (Sec.~\ref{subsec:clinical_application_deployment_proxy}), pooled AUROC is 0.687 [0.629, 0.740] for $H_R$ versus 0.648 [0.600, 0.701] for $H_{R|I}$ (not significant, $p = 0.20$). Crucially, $H_R$ cannot distinguish whether a high-uncertainty query needs clarification, model review, or both. CLUES is designed to \emph{decompose} uncertainty rather than maximize single-score prediction, providing diagnostic resolution unavailable from $H_R$ alone (Sec.~\ref{sec:regime_analysis}).

\subsection{Uncertainty Regime Analysis}
\label{sec:regime_analysis}

Given sparse per-condition errors, we pool all data for regime analysis. We partition queries using pooled median thresholds for $H_I$ and $H_{R|I}$, yielding a 2$\times$2 structure (Table~\ref{tab:regime_2x2}).

\begin{table}[htbp]
\centering
\setlength{\tabcolsep}{4pt}
\begin{tabular}{@{}lcccc@{}}
\toprule
\textbf{Regime} & \textbf{$H_I$} & \textbf{$H_{R|I}$} & \textbf{C/I} & \textbf{Err} \\
\midrule
I: Confident & Low & Low & 534/8 & 1.5\% \\
II: Ambiguity & High & Low & 530/18 & 3.3\% \\
III: Instability & Low & High & 529/21 & 3.8\% \\
IV: Compound & High & High & 492/48 & \textbf{8.9\%} \\
\midrule
\multicolumn{3}{@{}l}{\textbf{Total}} & 2085/95 & 4.4\% \\
\bottomrule
\end{tabular}
\caption{Uncertainty Regime Analysis. 2×2 decomposition by $H_I$ and $H_{R|I}$ median thresholds. C/I = Correct/Incorrect.}
\label{tab:regime_2x2}
\end{table}

\noindent\textbf{Error Gradient and Routing Implications.}
Error rates increase from Regime I (1.5\%) to Regime IV (8.9\%). Each regime contains $\sim$25\% of queries, yet Regime IV concentrates 51\% of all errors (48/95). A routing strategy sending only Regime IV to human review would examine one quarter of queries while catching half of all failures. Within-regime AUROC in Regime IV ($\approx$0.52 for all metrics) indicates that continuous entropy values provide limited additional signal once the regime is determined.
The decomposition's value is further demonstrated among high-uncertainty queries where both $H_I$ and $H_R$ are above median ($N=712$): these appear identical under conventional metrics, yet $H_{R|I}$ separates them into Regime A (4.3\% error) and Regime B (9.2\% error). This difference is significant (odds ratio OR = 2.24, $p < 0.02$) and holds across all five models using a single global threshold.

\section{Summary}
\label{sec:summary}

We introduced CLUES, a framework that decomposes semantic uncertainty into ambiguity ($H_I$) and conditional instability ($H_{R|I}$) via a Schur-complement construction over an interpretation--result bipartite graph. Across open-domain QA and clinical Text-to-SQL, $H_{R|I}$ is competitive with, and often improves upon, state-of-the-art Kernel Language Entropy~\cite{nikitin2024kernel} for failure prediction, while providing diagnostic resolution unavailable from a single score. Regime analysis based on $(H_I, H_{R|I})$ surfaces distinct failure patterns and maps them to targeted interventions: clarification for high ambiguity, human review for high instability. CLUES is model-agnostic and requires only output sampling; future work will explore adaptive routing and uncertainty propagation in agentic pipelines.

\section{Limitations}
\label{sec:limitations}
\noindent\textbf{Computational Cost.}
CLUES requires multiple LLM calls per query (interpretations $\times$ samples; 9 in our setup). Calls can be parallelized, though cost remains a consideration for large-scale deployment.

\noindent\textbf{Interpretation Quality.}
CLUES assumes generated interpretations meaningfully capture query ambiguity. If the disambiguation procedure produces trivial or redundant interpretations, $H_I$ may underestimate true input ambiguity, and the Schur complement may not isolate instability effectively. We observed strong results with frontier LLMs, but interpretation quality likely degrades with weaker models.

\noindent\textbf{Routing Efficiency in Sparse Error Regimes.}
In high-accuracy settings (95.6\% in deployment), regime-based routing catches 51\% of errors by reviewing 25\% of queries, twice the efficiency of random selection, though not yet sufficient for fully automated triage. The fundamental challenge is sparse errors: with only 95 failures ($\sim$4\% error rate), fine-grained uncertainty signals have limited discriminative power. Applications with higher base error rates may benefit more substantially from regime-based routing.

\section{Bibliographical References} 
\bibliographystyle{lrec2026-natbib}

\bibliography{custom,anthology-1,anthology-2}

\end{document}